\documentclass[a4paper,11pt]{article}
\usepackage{graphicx,amsmath,amssymb,latexsym,array,fullpage}

\newcommand{\expdif}[1]{D^{(#1)}}
\newcommand{\partf}{\ensuremath{\text{\rm Z}}}
\newcommand{\xspace}{\ensuremath{\mathcal X}}
\newcommand{\zspace}{\ensuremath{\mathcal Z}}
\newcommand{\regret}{\ensuremath{\widehat{\mathcal R}}}

\newcommand{\ceil}[1]{{\left\lceil#1\right\rceil}}
\newcommand{\floor}[1]{{\left\lfloor#1\right\rfloor}}
\newcommand{\var}{\textnormal{\rm var}}

\newcommand{\reals}{{\mathbb{R}}}

\newcommand{\nats}{{\mathbb{N}}}

\newcommand{\redundancy}{{\mathcal{R}}}
\newcommand{\model}{{\mathcal{M}}}

\newcommand{\di}{{\delta_i}}

\newcommand{\mstar}{{\mu^*}}
\newcommand{\mdagger}{{\mu^\dagger}}
\newcommand{\mihat}{{{\hat\mu}_i}}
\newcommand{\mnhat}{{{\hat\mu}_n}}
\newcommand{\mitilde}{{\tilde\mu}_i}

\newcommand{\tstar}{{\theta^*}}

\newcommand{\mmstar}{{M_\mstar}}

\newcommand{\mmihat}{{M_\mihat}}

\newcommand{\mmu}{{M_\mu}}

\newcommand{\mtheta}{{M_\theta}}
\newcommand{\mtstar}{{M_\tstar}}

\newcommand{\qed}{\hfill\ensuremath{\Box}}

\newtheorem{definition}{Definition}
\newtheorem{example}{Example}
\newtheorem{theorem}{Theorem}
\newtheorem{lemma}[theorem]{Lemma}
\newtheorem{proposition}[theorem]{Proposition}

\newtheorem{condition}{Condition}

\newenvironment{proof}[1][Proof]{\begin{trivlist}\item[\hskip \labelsep {\bfseries #1}]}{\qed\end{trivlist}}

\begin{document}
\title{Asymptotic Log-loss of Prequential Maximum Likelihood Codes}
\author{Peter Gr\"unwald and Steven de Rooij}
\maketitle

\begin{abstract} We analyze the Dawid-Rissanen \emph{prequential
    maximum likelihood codes\/} relative to one-\break parameter exponential
  family models $\model$. If data are i.i.d. according to an (essentially)
  {\em arbitrary\/} $P$, then the redundancy grows at rate ${1\over2}
  c \ln n$.  We show that $c = \sigma_1^2/ \sigma_2^2$, where
  $\sigma_1^2$ is the variance of $P$, and $\sigma_2^2$ is the
  variance of the distribution $M^*\in\model$ that is closest to $P$ in
  KL divergence.  This shows that prequential codes behave quite
  differently from other important universal codes such as the 2-part
  MDL, Shtarkov and Bayes codes, for which $c = 1$.  This behavior is
  undesirable in an MDL model selection setting.
\end{abstract}

\section{Introduction}\label{sec:introduction}
Universal coding lies at the basis of on-line prediction algorithms
for data compression and gambling purposes. It has been extensively
studied in the COLT community, typically under the name of `sequential
prediction with log loss', see, for example
\cite{Freund96,AzouryW01,CesaBianchiL01}.  It also underlies
Rissanen's theory of MDL (minimum description length) learning
\cite{BarronRY98,Grunwald05} and Dawid's theory of prequential model
assessment \cite{Dawid84}.  Roughly, a code is {\em universal\/} with
respect to a set of candidate codes ${\cal M}$ if it achieves small
{\em redundancy\/}: it allows one to encode data using not many more
bits than the optimal code in ${\cal M}$. The redundancy is very
closely related to the {\em expected regret}. which is perhaps more
widely known within the COLT community -- we compare the two notions
in Section~\ref{sec:regret}.  The main types of universal codes are
the {\em Shtarkov\/} or {\em NML\/} code, the {\em Bayesian mixture\/}
code, the {\em 2-part MDL code\/} and the {\em prequential maximum
  likelihood (ML) code}, also known as the `ML plug-in code' or the
`predictive MDL code' \cite{BarronRY98,Grunwald05}. This code was
introduced independently by Rissanen \cite{Rissanen84} in the context
of MDL learning and by Dawid \cite{Dawid84}, who proposed it as a
probability forecasting strategy rather than directly as a code. The
underlying ideas are explained in Section~\ref{sec:introres}. Here we
study the case where no code in $\model$ corresponds to the data
generating distribution $P$. We find that in this case, the redundancy
of the prequential code can be quite different from that of the other
three methods.  Specifically, if $\model$ is a one-dimensional
exponential family, then the redundancies are ${1 \over 2} c \ln n +
O(1)$.  Whereas it is known that for the Bayes, NML and 2-part codes,
under regularity conditions on $P$ and $\model$, we have $c=1$
(Section~\ref{sec:regret}), we determine $c$ for the prequential code
and find that, depending on properties of $P$ and $\model$, it can be
either larger or smaller than $1$.

\paragraph{Relevance}
Our result has at least three important consequences, which are
discussed further in Section~\ref{sec:consequences}.

\begin{description}
\item[1. Practical consequence for data compression] When prequential codes are used for data compression in the
  realistic situation that $P \not \in \model$, then depending on the
  situation they can behave either better or worse than the Bayesian
  and NML codes (end of Section~\ref{sec:introres}).
\item[2. Practical consequence for MDL learning/model selection] In the
  case of model selection between two
  nonoverlapping parametric models, our results suggest (but do not
  prove)  that the prequential plug-in codes typically behave
  worse (and never better) than the Bayesian or NML code. 
We have experimental evidence for this for the Poisson and geometric families.
\item[3. Theoretical] Our result implies that, under misspecification,
  the {\em Kullback-Leibler (KL) risk\/} of efficient estimators
  behaves in a fundamentally different way from the KL risk of
  estimators such as the Bayes predictive distribution which are not
  restricted to lie in the model $\model$ under consideration.
\end{description}

\paragraph{Contents}
The remainder of the paper is organized as follows. In
Section~\ref{sec:introres} we informally state and explain our result,
and we discuss how it relates to previous results.
Section~\ref{sec:main} contains the formal statement of our main
result (Theorem~\ref{thm:main}), as well as a brief proof sketch. We
show that a version of our result still holds if `redundancy' is
replaced by `expected regret' in Section~\ref{sec:regret}.  We discuss
further issues regarding our result in
Section~\ref{sec:variations}. We explain the relevance of our result,
including the consequences listed above, in
Section~\ref{sec:consequences}.  Section~\ref{sec:proof} proves our
main result. The proof makes use of several lemmas which are stated
and proven in Section~\ref{section:building_blocks}. The second
result, discussed in Section~\ref{sec:regret}, is proven in
Section~\ref{sec:proofb}. The paper ends with a conclusion.

\section{Main Result, Informally}
\label{sec:introres}
Suppose $\model = \{ \mtheta \; : \; \theta \in \Theta \}$ is a
$k$-dimensional parametric family of distributions, and $Z_1, Z_2,
\ldots$ are i.i.d. according to some distribution $P \in \model$. 
The {\em redundancy\/} of a universal code $U$ with respect to
$P$ is defined as
\begin{equation}\label{eq:red}
\redundancy_{U}(n) 
:= E_{P}[L_U(Z_1, \ldots, Z_n)] - 
\inf_{\theta \in \Theta} E_{P} [- \ln \mtheta(Z_1, \ldots, Z_n)],
\end{equation}
where $L_U$ is the length function of $U$ and $\mtheta(Z_1, \ldots,
Z_n)$ denotes the probability mass or density of $Z_1, \ldots, Z_n$
under distribution $\mtheta$; these and other notational conventions are
detailed in Section~\ref{sec:main}. By the information inequality 
\cite{CoverT91} the second term is minimized for $\mtheta = P$, so that
\begin{equation}\label{eq:red2}
\redundancy_{U}(n) 
= E_{P}[L_U(Z_1, \ldots, Z_n)] - 
E_{P}[- \ln P(Z_1, \ldots, Z_n)],
\end{equation}
Thus, (\ref{eq:red2}) can be interpreted as the expected number of
additional nats one needs to encode $n$ outcomes if one uses the code
$U$ instead of the optimal (Shannon-Fano) code with lengths $- \ln
P(Z_1, \ldots, Z_n)$. A good universal code achieves small redundancy
for all or `most' $P \in \model$ (the relation to the concept of
`regret' is discussed in Section~\ref{sec:regret}).

The four major types of universal codes, Bayes, NML, 2-part and
prequential ML, all achieve redundancies that are (in an appropriate
sense) close to optimal. Specifically, under regularity conditions on
$\model$ and its parameterization, these four types of universal codes
all satisfy, for all $P \in \model$,

\begin{equation}
\label{eq:bic}
\redundancy(n) = \frac{k}{2} \ln n + O(1),
\end{equation}
where the $O(1)$ may depend on $\theta$ and the universal code
used. (\ref{eq:bic}) is the famous `$k$ over $2$ log $n$ formula',
refinements of which lie at the basis of most practical
approximations to MDL learning, see \cite{Grunwald05}. 

In this paper we consider the case where the data are i.i.d.
according to an {\em arbitrary\/} $P$ not necessarily in the model
$\model$. To emphasize that the redundancy is measured relative to the
element of the model that minimizes the codelength rather than to $P$,
we use the term \emph{relative redundancy} rather than just
\emph{redundancy}. Its definition (\ref{eq:red}) remains unchanged,
but it can no longer be rewritten as (\ref{eq:red2}): Assuming it
exists and is unique, let $\mtstar$ be the element of $\model$ that
minimizes KL divergence to $P$:
$$\tstar := \arg \min_{{\theta} \in \Theta} D(P \| M_{\theta}) = 
\arg \min_{{\theta} \in \Theta} E_{P} [ - \ln \mtheta(Z)],$$
where the equality follows from the definition of the KL divergence
  \cite{CoverT91}. Then the relative redundancy satisfies
\begin{equation}
\label{eq:redb}
\redundancy_{U}(n) = E_{P}[L_U(Z_1, \ldots, Z_n)] - 
E_{P}[- \ln \mtstar(Z_1, \ldots, Z_n)]. 
\end{equation}
It turns out that for the NML, 2-part MDL and Bayes codes, the
relative redundancy (\ref{eq:redb}) with $P \not \in \model$, still
satisfies (\ref{eq:bic}), at least under conditions on $\model$ and
$P$; see Section~\ref{sec:regret}. In this paper, we show for the
first time that (\ref{eq:bic}) does {\em not\/} hold for the
prequential ML code. The prequential ML code $U$ works by sequentially
predicting $Z_{i+1}$ using a (slightly modified) ML or Bayesian MAP
estimator $\hat{\theta}_i = \hat{\theta}(z^i)$ based on the past data,
that is, the first $i$ outcomes $z^i = z_1, \ldots, z_i$. The total
codelength $L_U(z^n)$ on a sequence $z^n$ is given by the sum of the
individual `predictive' codelengths (log losses): $L_U(z^n) =
\sum_{i=0}^{n-1} [ - \ln M_{\hat{\theta}_i}(z_{i+1})]$.  In our main
theorem, we show that if $L_U$ denotes the prequential ML code length,
and $\model$ is a regular one-parameter exponential family $(k=1)$,
then

\begin{equation}
\label{eq:redc}
{\redundancy}_{U}(n) = \frac{1}{2} \frac{\var_P
X}{\var_{M_{{\theta}^*}} X } \ln n + O(1),
\end{equation}
where $X$ is the sufficient statistic of the family. In
Example~\ref{ex:ample} below we give an example of the phenomenon. The
result holds as long as $\model$ and $P$ satisfy
Condition~\ref{condition} defined below. Essentially, as long as the
fourth moment of $P$ exists, the condition holds for all exponential
families we checked, including the Poisson, geometric, exponential,
normal with fixed mean or variance and Pareto distributions.  The
result indicates that the redundancy can be both larger and smaller
than ${1 \over 2} \ln n$, depending on the variance of the `true' $P$.
We can only guarantee that the two variances are the same if $P \in
\model$, in which case $\mtstar = P$. It immediately follows that in
practical data compression tasks, whenever $P \not \in \model$, the
redundancy of the prequential ML code can be both smaller and larger
than that of the Bayesian code, depending on the situation. This is
the first of the three implications of our result, listed in
Section~\ref{sec:introduction}. We postpone discussion of the other
two implications to Section~\ref{sec:consequences}.

\begin{example}
\label{ex:ample}
\rm Let $\model$ be the family of Poisson distributions, parameterized
by their mean $\mu$.  Then the ML estimator $\hat{\mu}_i$ is the
empirical mean of $z_1, \ldots, z_i$. Suppose $Z$, $Z_1$, $Z_2$,
$\ldots$ are i.i.d. according to a degenerate $P$ with $P(Z = 4) = 1$.
Since the sample average is a sufficient statistic for the Poisson family,
$\mihat$ will be equal to $4$ for all $i\ge1$. On the other hand,
$\mu^*$, the parameter (mean) of the distribution in $\model$ closest
to $P$ in KL-divergence, will be equal to $4$ as well.  Thus the
redundancy (\ref{eq:redb}) of the prequential ML code is given by
\begin{eqnarray}
\redundancy_U(n) = 
\sum_{i=0}^{n-1} [ - \ln \mmihat(4) + \ln \mmstar(4)]& = &
- \ln M_{\hat{\mu}_0}(4) + \ln M_4(4) +
\sum_{i=1}^{n-1} [ - \ln M_4(4) + \ln M_4(4)]\nonumber\\
& = &
- \ln M_{\hat{\mu}_0}(4) + \ln M_4(4) = O(1),
\end{eqnarray}
assuming an appropriate definition of $\hat{\mu}_0$. In the case of
the Poisson family, the outcome $Z$ is equal to the sufficient
statistic $X$ in (\ref{eq:redc}).  Since $\var_P Z = 0$, this example
agrees with (\ref{eq:redc}).
\end{example}

\paragraph{Related Work}
There are a plethora of results concerning the redundancy and/or the
regret for the prequential ML code, for a large variety of models
including multivariate exponential families, ARMA processes,
regression models and so on. Examples are
\cite{Rissanen1986c,Gerencser1987,Hemerly1989a,Wei1990,LiY00}. In all
these papers it is shown that either the regret or the redundancy
grows as ${k\over2}\ln n$ + $o(\ln n)$, either in expectation or
almost surely. \cite{LiY00} even evaluates the remainder term
explicitly. The reason that these results do not contradict ours, is
that in all these papers, one studies the case where the generating
distribution $P$ is in the model, in which case automatically
$\var_{M^*}(X)=\var_P(X)$. In other cases \cite{Freund96,Rissanen86b},
regret of a prequential ML-type code is evaluated on an individual
sequence basis, and it is found that the regret grows as ${k\over2}\ln
n + O(1)$ for all sequences whose ML estimator remains bounded away
from the boundary of the space. The reason that these results do not
contradict ours, is that in all cases that have been examined (and
that we know of), the model is {\em complete}, i.e. it contains {\em
  all\/} distributions that can be defined on the sample space for 1
outcome. Then, if data are i.i.d. according to some $P$, $P$ {\em
  must\/} be in $\model$, and we automatically get
$\var_{M^*}(X)=\var_P(X)$. An example is \cite{Freund96} which uses
the Bernoulli model.  Apparently, we are the first to study the
redundancy and regret for incomplete models under general
circumstances.
\section{Main Result, Formally}
\label{sec:main}
In this section, we introduce our notation, we define our quantities
of interest, we state our main result and we give a short idea of the
proof. The complete proof is given in
Sections~\ref{sec:proof}-\ref{section:building_blocks}. 

\paragraph{Notational Conventions}
Throughout this text we use nats rather than bits as units of
information.  Outcomes are capitalized if they are to be interpreted
as random variables instead of instantiated values. A sequence of
outcomes $z_1,\ldots, z_n$ is abbreviated to $z^n$. We write $E_{P}$
as a shorthand for $E_{Z \sim P}$, the expectation of $Z$ under
distribution $P$. When we consider a sequence of $n$ outcomes
independently distributed $\sim P$, we use $E_{P}$ even as a shorthand
for the expectation of $(Z_1, \ldots, Z_n)$ under the $n$-fold product
distribution of $P$.  Finally, $P(Z)$ denotes the probability mass
function of $P$ in case $Z$ is discrete-valued, and it denotes the
density of $P$, in case $Z$ takes its value in a continuum. When we
write `density function of $Z$', then, if $Z$ is discrete-valued, this
should be read as `probability mass function of $Z$'. Note however
that in our main result, Theorem~\ref{thm:main} below, we do not
assume that the data generating distribution $P$ admits a density.

\paragraph{Exponential Families}
Let $\zspace$ be a set of outcomes, taking values either in a finite
or countable set, or in a subset of $k$-dimensional Euclidean space
for some $k \geq 1$. Let $X: \zspace \rightarrow \reals$ be a random
variable on $\zspace$, and let $\xspace = \{ x \in \reals \; : \;
\exists z \in \zspace: X(z) = x\}$ be the range of $X$.

Exponential family models are families of distributions on $\zspace$
defined relative to a random variable $X$ (called `sufficient
statistic') as defined above, and a function $h: \zspace \rightarrow
[0,\infty)$. We let $\partf(\eta) :=\int_{z\in\zspace}e^{-\eta
X(z)}h(z)dz$ (where the integral is to be replaced by a sum for
countable $\zspace$), and we let $\Theta_\eta
:=\{\eta\in\reals:\partf(\eta)<\infty\}$.

\begin{definition}[Exponential family]\label{def:expfam}
The \emph{single parameter exponential family} {\rm \cite{kassvos}}
with {\em sufficient statistic\/} $X$ and {\em carrier\/} $h$ is the
family of distributions with densities $M_\eta(z) :={1\over
\partf(\eta)}e^{-\eta X(z)}h(z)$, where $\eta\in
\Theta_\eta$. $\Theta_\eta$ is called the \emph{natural parameter
space}. The family is called \emph{regular} if $\Theta_\eta$ is an
open interval of $\reals$.
\end{definition}
In the remainder of this text we only consider single parameter,
regular exponential families where the mapping from $\Theta_\eta$ to the corresponding set of distributions is 1-to-1, but these qualifications will henceforth
be omitted. Examples of this wide family of models include the
Poisson, geometric and multinomial families, and the model of all
Gaussian (normal) distributions with a fixed variance, or with a fixed
mean. In the first four cases, we can take $X$ to be the identity, so
that $X = Z$ and $\xspace =
\zspace$. In the case of the normal family with fixed mean, $\sigma^2$ 
becomes the sufficient statistic and we have $\zspace =
\reals$, $\xspace =[0, \infty)$ and $X = Z^2$. 

The statistic $X(z)$ is sufficient for $\eta$ \cite{kassvos}. This
suggests reparameterizing the distribution by the expected value of
$X$, which is called the \emph{mean value parameterization}. The
function $\mu(\eta)=E_{M_\eta}[X]$ maps parameters in the natural
parameterization to the mean value parameterization. It is a
diffeomorphism (it is one-to-one, onto, infinitely often
differentiable and has an infinitely often differentiable inverse)
\cite{kassvos}. Therefore the mean value parameter space $\Theta_\mu$
is also an open interval of $\reals$. We note that for some models
(such as Bernoulli and Poisson), the parameter space is usually given
in terms of the a non-open set of mean-values (e.g., $[0,1]$ in the
Bernoulli case). In this case, to make the model a regular exponential
family, we have to restrict the set of parameters to its own interior.
Henceforth, whenever we refer to a standard statistical model such as
Bernoulli or Poisson, we assume that the parameter set has been
restricted in this sense.

We are now ready to define the prequential ML model. This is a
distribution on infinite sequences $z_1, z_2, \ldots \in
\zspace^{\infty}$, recursively defined 
in terms of the  distributions of $Z_{n+1}$ conditioned on 
$Z^n = z^n$, for all $n = 1,2 , \ldots$, all 
$z^n = (z_1, \ldots, z_n) \in \zspace^n$. In the definition, we use
the notation $x_i := X(z_i)$.

\begin{definition}[Prequential ML model]
\label{def:preq}
Let $\Theta_\mu$ be the mean value parameter domain of an exponential
family $\model=\{\mmu\mid\mu\in \Theta_\mu \}$.  Given $\model$ and
constants $x_0\in \Theta_{\mu}$ and $n_0>0$, we define the \emph{prequential ML
model} $U$ by setting, for all $n$, all $z^{n+1} \in \zspace^{n+1}$:
$$
   U(z_{n+1} \mid z^n) = M_{\hat{\mu}(z^n)}(z_{n+1}),
$$
where $U(z_{n+1} \mid z^n)$ is the density/mass function of $z_{n+1}$ conditional on $Z^n = z^n$, 
$$
    \hat{\mu}(z^n) :=\frac{x_0 \cdot n_0+\sum_{i=1}^n x_i}{n+n_0},
$$
and $M_{\hat{\mu}(z^n)}(\cdot)$ is the density of the distribution in $\model$ with mean $\hat{\mu}(z^n)$.
\end{definition}
We henceforth abbreviate $ \hat{\mu}(z^n)$ to $\mnhat$. 
We usually refer to the prequential ML model in terms of the
corresponding codelength function 
$$
L_U(z^n) = \sum_{i=0}^{n-1}  L_U(z_{i+1} \mid z_i) = 
\sum_{i=0}^{n-1} - \ln M_{\hat{\mu}_{i}}(z_{i+1}).
$$ To understand this definition, note that for exponential families,
for any sequence of data, the ordinary maximum likelihood parameter is
given by the average $n^{-1} \sum x_i$ of the observed values of $X$
\cite{kassvos}.  Here we define our prequential model in terms of a
slightly modified maximum likelihood estimator that introduces a `fake
initial outcome' $x_0$ with multiplicity $n_0$ in order to avoid
infinite code lengths (see the quote by Rissanen on ``inherent
singularity'' in Section~\ref{sec:consequences}) and to ensure that
the prequential ML code length of the first outcome is
well-defined. In practice we can take $n_0 = 1$ but our result holds
for any $n_0 > 0$. This definition can be reconciled with settings in
which the startup problem is resolved by ignoring the first few
outcomes, by setting $x_0$ to the ML estimator for the ignored
outcomes and $n_0$ to their number. It also allows our results to be
generalized to a number of other point estimators as discussed in
Section~\ref{sec:other}.

\bigskip\noindent% MAIN RESULT
With all our definitions in place we can state
our main result. 
\begin{theorem}[Main result]
\label{thm:main}
  Let $X, X_1, X_2, \ldots$ be i.i.d. $\sim P$, with $E_P[X] = \mstar$. 
Let  $\model$ be a single parameter exponential family with sufficient
statistic $X$ and $\mstar$ an
  element of the mean value parameter space. Finally let $U$ denote
  the prequential ML model with respect to $\model$.  If $\model$ and $P$
  satisfy Condition 1 below, then
$$
\redundancy_U(n)={\var_P X \over\var_\mmstar X}{1\over2}\ln n + O(1).
$$
\end{theorem}
To reconcile this with the informal statement (\ref{eq:redc}), notice
that $M_{\mu^*}$ is the element of $\model$ achieving the smallest
expected codelength, i.e. it achieves $\inf_{\mu \in \Theta_\mu} D(P \|
M_{\mu})$ \cite{kassvos}.

\begin{condition}\label{condition}
We require that the following holds both for $T:= X$ and $T:=- X$:
\begin{itemize}
\item If $T$ is unbounded from above then there is a
  $k\in\{4,6,\ldots\}$ such that the first $k$ moments of $T$ exist under $P$
  and that ${d^4\over d\mu^4}D(\mmstar\|\mmu)=O\left(\mu^{k-6}\right)$.
\item If $T$ is bounded from above by a constant $g$ then ${d^4\over
d\mu^4}D(\mmstar\|\mmu)$ is polynomial in $1/(g-\mu)$.
\end{itemize}
\end{condition}
The condition implies that Theorem~\ref{thm:main} can be applied to
most single-parameter exponential families that are relevant in
practice. To illustrate, we have computed the fourth derivative of the
divergence for a number of exponential families; all parameters beside
the mean are treated as fixed values. The results are listed in
Figure~\ref{fig:modlist}. As can be seen from the figure, for these
exponential families, our condition applies whenever the fourth moment
of $P$ exists. Note in particular that the condition requires $\var_P
X < \infty$.

\begin{figure}[t]
\begin{center}
\newcommand{\tabline}[3]{#1&$\displaystyle{#2}$&$\displaystyle{#3}$\\}
\setlength{\extrarowheight}{12pt}
\begin{tabular}{l|c|c|}
\tabline{}{M_{\mu^*}(x)}{{d^4\over d\mu^4}D(\mmstar\|\mmu)}
\hline
\tabline{Bernoulli}{(\mstar)^x(1-\mstar)^{(1-x)}}{{6\mstar\over\mu^4}+{6(1-\mstar)\over(1-\mu)^4}}
\tabline{Poisson}{e^\mstar\mstar^x\over x!}{6\mstar\over\mu^4}
\tabline{Geometric}{\theta^x(1-\theta)={(\mstar)^x\over(\mstar+1)^{x+1}}}{{6\mstar\over\mu^4}-{6(\mstar+1)\over(\mu+1)^4}}
\tabline{Exponential}{{1\over\mstar}e^{-x/\mstar}}{-{6\over\mu^4}+{24\mstar\over\mu^5}}
\tabline{Normal (fixed mean $=0$)}{{1\over\sqrt{2\pi\mstar
      x}}e^{-{x\over 2\mstar}}}{-{3\over\mu^4}+{12\mstar\over\mu^5}}
%steven: changed \mu to \mstar in the following. By the way, I still don't
%see how it is an exponential family.
\tabline{Normal (fixed variance $=1$)}{{1\over\sqrt{2\pi}}e^{-{1 \over
      2}(x-\mstar)^2}}{0}
\tabline{Pareto}{{ab^a\over x^{a+1}}\hbox{~for~}b={a-1\over
    a}\mstar}{6a\over\mu^4}
\hline
\end{tabular}
\caption{\label{fig:modlist}${d^4\over d\mu^4}D(\mmstar\|\mmu)$ for a number of
  exponential families. For the normal distribution we use mean $0$,
  and we list a reparametrization of the density function such that
  the density of the squared outcomes is given as a function of the
  variance, which is confusingly but correctly called $\mstar$ 
  %steven: changed another \mu to \mstar here.
  here: the random variable $X$ in Theorem~\ref{thm:main} is really
  the observed value of $z^2$ rather than $z$ itself, so that its mean
  is $E[X] = E[Z^2]$, which is the variance of the normal
  distribution.}
\end{center}
\end{figure}
The reason why we need Condition~\ref{condition} is best explained by
sketching the proof of Theorem~\ref{thm:main}:

\paragraph{Brief Proof Sketch}
The precise proof of Theorem~\ref{thm:main}, given in
Section~\ref{sec:proof}, is very technical. Here we merely describe
the underlying ideas, which are relatively simple. Consider first the case
$n_0 = 0$, so that for $n \geq 1$, $\hat{\mu}_n$ is just the standard
ML estimator. Let $z^i$ be the initial $i$ outcomes of an arbitrary
sequence $z^n = z_1, z_2, \ldots, z_n$. As is well-known, a
straightforward second-order Taylor expansion of $D(M_{\mstar} \|
M_{\hat{\mu}(z^i)})$ around $\mu^*$ gives
%steven: changed \hat\mu to \mihat in D(.||.) (the index was missing)
\begin{equation}
\label{eq:rem}
D(\mmstar \| \mmihat) =
\frac{1}{2} I(\mu^*) (\mihat - \mstar)^2  + \mbox{Remainder}.
\end{equation}
Here $I(\mu^*)$ is the Fisher information in one observation,
evaluated at $\mu^*$, see Section~\ref{sec:proof}. 
For exponential families in their mean-value parameterization, another
standard result \cite{kassvos} says that for all $\mu$,

\begin{equation}\label{eq:efficient}
I(\mu) = \frac{1}{\var_{M_\mu} X}
\end{equation}
Therefore, ignoring the remainder term and the term for $i=0$, we get 
\begin{multline}
\label{eq:hap}
\sum_{i=0}^{n-1}\,\mathop E_{\mihat\sim
  P}\left[D(\mmstar\| \mmihat)\right] \approx
\frac{1}{2} \sum_{i=1}^{n-1}
\frac{E_{P}(\hat{\mu}_i - \mstar)^2}{\var_{M_{\mu^*}} X}  = \\
\frac{1}{2} \frac{\var_P X}{\var_{M_{\mu^*}} X} \sum_{i=1}^{n-1} \frac{1}{i}
= 
\frac{1}{2} \frac{\var_P X}{\var_{M_{\mu^*}} X} \ln n + O(1),
\end{multline}
Here the first approximate equality follows by (\ref{eq:rem}) and
(\ref{eq:efficient}). The second follows because for exponential
families, the ML estimator $\hat{\mu}_i$ is just the empirical average
$i^{-1} \sum x_i$, so that $E_{P}(\hat{\mu}_i - \mstar)^2 = \var
(i^{-1} \sum_{j=1}^i X_j) = i^{-1} \var X$.  
Thus, Theorem~\ref{thm:main} follows if we can
show (a) that the left-hand side of (\ref{eq:hap}) is equal to the
relative redundancy $\redundancy_U(n)$ and (b) that, as $n \rightarrow
\infty$, the remainder terms in (\ref{eq:rem}), summed over $n$ as in
(\ref{eq:hap}), form a convergent series (i.e. sum to something
finite). Result (a) follows relatively easily by rewriting the sum
using the chain rule for relative entropy and using the fact that $X$
is a sufficient statistic (Lemma~\ref{lem:redundancy}). The truly
difficult part of the proof is (b), shown in Lemma~\ref{lem:hard}. It
involves infinite sums of expectations over unbounded fourth-order
derivatives of the KL divergence.  To make this work, we (1) slightly
modify the ML estimator by introducing the initial fake outcome $x_0$.
And (2), we need to impose Condition~\ref{condition}.  To understand
it, consider the case $T = X$, $X$ unbounded from above.  The
condition essentially expresses that, as $\hat{\mu}$ increases to
infinity, the fourth order Taylor-term does not grow too
fast. Similarly, if $X$ is bounded from above by $g$, the condition
ensures that the fourth-order term grows slowly enough as $\hat{\mu}
\uparrow g$. The same requirements are imposed for decreasing
$\hat{\mu}$.

\section{Redundancy vs. Regret}  
\label{sec:regret}
The `goodness' of a universal code
relative to a model $\model$ can be measured in several ways: rather
than using redundancy (as we did here), one can also choose to measure
codelength differences in terms of {\em regret}, where one has a
further choice between {\em expected regret\/} and {\em worst-case
  regret\/} \cite{BarronRY98}. Here we only discuss the implications
of our result for the expected regret measure.

Let $\model = \{ M_{\theta} \mid \theta \in \Theta \}$ 
be a family of distributions parameterized  by $\Theta$.
Given a sequence $z^n = z_1, \ldots, z_n$ and a universal code $U$ for
$\model$ with lengths
$L_U$, the {\em regret} of $U$ on sequence $z^n$ is defined as
\begin{equation}
\label{eq:regret}
L_U(z^n) - \inf_{\theta \in \Theta} [ - \ln M_{\theta}(z^n) ].
\end{equation}
Note that if the (unmodified) ML estimator $\hat{\theta}(z^n)$ exists,
then this is equal to $L_U(z^n) + \ln M_{\hat{\theta}(z^n)}(z^n)$.
Thus, one compares the codelength achieved on $z^n$ by $U$ to
the best possible that could have been achieved on that particular
$z^n$, 
using any of
the distributions in $\model$. Assuming $Z_1, Z_2, \ldots$ are
i.i.d. according to some (arbitrary) $P$, one 
may now consider the expected
regret
$$
\regret_U(n) := E_{P} [L_U(Z^n) -  \inf_{\theta \in \Theta} [
- \ln M_{\theta}(Z^n)]],
$$
To quantify the 
difference between redundancy and expected regret, consider the
function 
$$
d(n) := \inf_{\theta \in \Theta} 
E_{P} [- \ln M_{\theta}(Z^n)] - 
E_{P} \; [ \inf_{\theta \in \Theta} [- \ln M_{{\theta}}(Z^n)] ], 
$$
and note that for any universal code, $\redundancy_U(n) 
- \regret_U(n) = d(n)$. 
In case $P \in \model$, then under regularity conditions on
$\model$ and its parameterization, it can be shown \cite{ClarkeB90} that 
\begin{equation}
\label{eq:alleen}
\lim_{n \rightarrow \infty} d(n) = \frac{k}{2},
\end{equation}
where $k$ is the dimension of $\model$.
In our case, where $P$ is not necessarily in $\model$, we
have the following:
\begin{theorem}
\label{thm:regret}
Let $\xspace$ be finite. Let $P$, $M_{\mu}$ and $\mu^*$  be as in
Theorem~\ref{thm:main}.
Then
\begin{equation}
\lim_{n \rightarrow \infty} d(n) = \frac{1}{2} \frac{\var_{P}
  X}{\var_{M_{\mu^*}} X}.
\end{equation}
\end{theorem}
Once we are dealing with $1$-parameter families, in the special case
that $P \in \model$, this result reduces to (\ref{eq:alleen}).  We
conjecture that, under a condition similar to
Condition~\ref{condition}, the same result still holds for general,
not necessarily finite or countable or bounded $\xspace$, but at the
time of writing this submission we did not yet find the time to sort
out the details. In any case, our result is sufficient to show that in
some cases (namely, if $\xspace$ is finite), we have
$$
\regret_U(n) = \frac{1}{2} \frac{\var_{P}
  X}{\var_{M_{\mu^*}} X} \ln n + O(1),
$$
so that, up to $O(1)$-terms, the redundancy and the regret of the
prequential ML code behave in the same way. 

Incidentally, Theorem~\ref{thm:regret} can be used to substantiate the
claim we made in Section~\ref{sec:introres}, which stated that the Bayes
(equipped with a strictly positive differentiable prior), NML and
2-part codes still achieve relative redundancy of ${1 \over 2} \ln n$
if $P \neq \model$, at least if $\xspace$ is finite. Let us informally
explain why this is the case.  It is easy to show that Bayes, NML and
(suitably defined) 2-part codes achieve regret ${1 \over 2} \ln n +
O(1)$ for {\em all\/} sequences $z_1, z_2, \ldots$ such that
$\hat{\theta}(z^n)$ is bounded away from the boundary of the parameter
space $M$, for all large $n$ \cite{BarronRY98,Grunwald05}. It then
follows using, for example, the Chernoff bound that these codes must
also achieve expected regret ${1 \over 2} \ln n + O(1)$ for {\em
all\/} distributions $P$ on $\xspace$ that satisfy $E_P[X] = \mu^* \in
\Theta_\mu$.  Theorem~\ref{thm:regret} then shows that they also
achieve relative redundancy ${1 \over 2} \ln n + O(1)$ for {\em all\/}
distributions $P$ on $\xspace$ that satisfy $E_P[X] = \mu^* \in
\Theta_\mu$. We omit further details.

\section{Variations of Prequential Coding}\label{sec:variations}

\subsection{Justifying Our Modification of the ML Estimator}
If the prequential code is based on the ordinary ML estimator ($n_0 =
0$ in Definition~\ref{def:preq}) then, apart from being undefined for
the first outcome, it may achieve infinite codelengths on the observed
data. A simple example is the Bernoulli model. If we first observe
$z_1 = 0$ and then $z_2 = 1$, the codelength of $z_2$ according to the
ordinary ML estimator of $z_2$ given $z_1$ would be $- \ln
M_{\hat{\mu}}(z_1)(z_2) = - \ln 0 = \infty$. There are several ways to
resolve this problem. We choose to add an `initial fake
outcome'. Another possibility that has been suggested (e.g.,
\cite{Dawid84}) is to use the ordinary ML estimator, but only start
using after having observed $m$ examples, where $m$ is the smallest
number such that $- \ln M_{\hat{\mu}(z^m)}(Z_{m+1})$ is guaranteed to
be finite, no matter what value $Z_{m+1}$ is realized.  The first $m$
outcomes may then be encoded by repeatedly using some code $L_0$ on
outcomes of $\zspace$, so that for $i \leq m$, the codelength of $z_i$
does not depend on the outcomes $z^{i-1}$. In the Bernoulli example,
one could for example use the code corresponding to $P(Z_i = 1) =
1/2$, until and including the first $i$ such that $z^i$ includes both
a $0$ and a $1$. Then it takes $i$ bits to encode the first $z^i$
outcomes, no matter what they are.  After that, one uses the
prequential code with the standard ML estimator.  It is easy to see
(by slight modification of the proof) that our theorem still holds for
this variation of prequential coding.  Thus, our particular choice for
resolving the startup problem is not crucial to obtain our result.
The advantage of our solution is that, as we now show, it allows us to
interpret our modified ML estimator also as a Bayesian MAP and
Bayesian mean estimator, thereby showing that the same behavior can be
expected for such estimators.

\subsection{Prequential Models with Other Estimators}\label{sec:other}

\paragraph{The Bayesian MAP estimator}
If a conjugate prior is used, the Bayesian maximum a-posteriori
estimator can always be interpreted as an ML estimator based on the
sample and some additional `fake data' (\cite{Berger85}; see also the
notion of {\em ESS (Equivalent Sample Size) Priors\/} discussed in,
for example, \cite{KontkanenMSTG00}).  Therefore, the prequential ML
model as defined above can also be interpreted as a prequential MAP
model for that class of priors, and the whole analysis carries over to
that setting.

\paragraph{The Bayesian mean estimator}
It follows by the work of Hartigan \cite[Chapter 7]{Hartigan83} on the
so-called `maximum likelihood prior', that by slightly modifying
conjugate priors, we can construct priors such that the Bayesian {\em
mean\/} rather than MAP estimator is of the form of our modified ML
estimator.

\paragraph{A Conjecture}
In some special cases, for example, the Bernoulli model, the
exponential family $\model$ covers all distributions that can be
defined on $\xspace$. In such cases, there exists no distribution with
mean $\mu^*$ and variance not equal to $\var_{M_{\mu^*}} X$, and the
${1 \over 2} \ln n + O(1)$ redundancy can always be achieved. But in
all other cases, three very reasonable and efficient \cite{Rice95}
estimators (ML, Bayes MAP, Bayes mean for a large class of reasonable
priors) cannot achieve ${1 \over 2} \ln n + O(1)$ in all
circumstances. This suggests that {\em no matter what
  in-model\footnote{See Section~\ref{sec:consequences}.}  estimator is
  used}, the prequential model cannot yield a relative redundancy of
${1\over2}\ln n$ independently of the variance of the data generating
distribution $P$.

\subsection{Rissanen's Predictive MDL Approach}
\label{sec:rissanen}
The MDL model selection criterion that is based on comparing the
prequential ML codelengths for the models under consideration is
called the \emph{Predictive MDL (PMDL)\/} criterion by Rissanen
\cite{Rissanen1986c}. It is closely related to the \emph{Predictive
Least Squares (PLS)\/} criterion \cite{Rissanen1989}
for regression models; PMDL can be seen as an MDL
justification for it. There has been some discussion on how to use
PMDL when the data are not ordered. The prequential ML codelength then
becomes redundant: the same data can be coded in any order, yielding
different code words. Rissanen suggests in \cite{Rissanen1986b} to use
the permutation of the outcomes that minimizes the codelength. Under
such a regime, Theorem~\ref{thm:main} is no longer applicable (since
the outcomes are no longer i.i.d.); however Example~\ref{ex:ample}
illustrates that circumstances in which the prequential ML codelength
and the NML codelength behave very differently remain, under {\em any\/}
regime that amounts to reordering the sample, including the one
suggested by Rissanen.
\section{Consequences}
\label{sec:consequences}
Why are these results interesting? We listed three significant
implications in Section~\ref{sec:introduction}, the introduction to
this paper. The first was evident from Theorem~\ref{thm:main}. Let us
now discuss the second and third in more detail.

\paragraph{Practical significance for Model Selection}
There exist a plethora of results showing that in various contexts, if
$P \in \model$, then the prequential ML code achieves optimal redundancy
(see Section~\ref{sec:introres}, Related Work). These strongly suggest
that it is a very good alternative for (or at least approximation to)
the NML or Bayesian codes in MDL model selection. Indeed, quoting
Rissanen \cite{Rissanen1989}:
\begin{quotation}
  ``If the encoder does not look ahead but instead computes the best
  parameter values from the past string, only, using an algorithm
  which the decoder knows, then no preamble is needed. The result is a
  \emph{predictive} coding process, one which is quite different from
  the sum or integral formula in the stochastic
  complexity.\footnote{The stochastic complexity is the codelength of
    the data $z_1, \ldots, z_n$ that can be achieved using the
    NML code.} And it is only because of a certain inherent
  singularity in the process, as well as the somewhat restrictive
  requirement that the data must be ordered, that we do not consider
  the resulting predictive code length to provide another competing
  definition for the stochastic complexity, but rather regard it as an
  approximation.''
\end{quotation}
Our result however shows that the prequential ML code may behave quite
differently from the NML and Bayes codes, thereby strengthening the
conclusion that it should not be taken as a definition of stochastic
complexity. Although there is only a significant difference if data
are distributed according to some $P \not \in \model$, the difference
is nevertheless very relevant in an MDL model selection context with
nonoverlapping models, even if one of the models under consideration
{\em does\/} contain the `true' $P$. To see this, suppose we are
comparing two models $\model_1$ and $\model_2$ for the same data, and
in fact, $P \in \model_1 \cup \model_2$.  For concreteness, assume
$\model_1$ is the Poisson family and $\model_2$ is the geometric
family. We want to decide which of these two models best explains the
data. According to the MDL Principle, we should associate with each
model a universal code (preferably the NML code). We should then pick
the model such that the corresponding universal codelength of the data
is minimized.  Now suppose we use the prequential ML codelengths
rather than the NML codelengths. Without loss of generality suppose
that $P \in \model_1$.  Then $P \not \in \model_2$. This means that
the codelength relative to $\model_1$ behaves essentially like the NML
codelength, but the codelength relative to $\model_2$ behaves
differently -- at least as long as the variances do not match (which
for example, is forcibly the case if $\model_1$ is Poisson and
$\model_2$ is geometric).  This introduces a bias in the model
selection scheme. We have found experimentally \cite{uai} that the
error rate for model selection based on the prequential ML code
decreases more slowly than when other universal codes are used. Even
though in some cases the redundancy grows \emph{more slowly} than
${1\over2}\ln n$, so that the prequential ML code is in a sense more
efficient than the NML code, model selection based on the prequential
ML codes behaves worse than Bayesian and NML-based model selection. We
provide a theoretical explanation for this phenomenon in \cite{uai}.
The practical relevance of this phenomenon stems from the fact that
the prequential ML codelengths are often a lot easier to compute than
the Bayes or NML codes, so that they are often used in applications
\cite{ModhaM98,KontkanenMT01}.
\paragraph{Theoretical Significance}
The result is also of theoretical-statistical interest: our theorem
can be re-interpreted as establishing bounds on the asymptotic {\em
Kullback-Leibler risk\/} of density estimation using ML and Bayes
estimators under misspecification $(P \not \in \model)$.  Our result
implies that, under misspecification, the KL risk of estimators such
as ML, which are required to lie in the model $\model$, behaves in a
fundamentally different way from the KL risk of estimators such as the
Bayes predictive distribution, which are not restricted to lie in
$\model$. Namely, we can think of every universal model $U$ defined as
a random process on infinite sequences as an {\em estimator\/} in the
following way: define, for all $n$,
$$\breve{P}_n := \Pr_U(Z_{n+1} = \cdot \mid Z_1 = z_1, \ldots, Z_n =
z_n),$$
a function of the sample $z_1, \ldots, z_n$. $\breve{P}_n$ can be
thought of as the `estimate of the true data generating distribution
upon observing $z_1, \ldots, z_n$'. In case $U$ is the prequential ML
model, $\breve{P}_n = M_{\hat{\theta}_n}$ is simply our modified ML
estimator. It is now important to note that for other universal
models, $\breve{P}_n$ is not required to lie in $\model$. An example
is the Bayesian universal code defined relative to some prior
$w$. This code has lengths $L'(z^n) := - \ln \int M_{\mu}(z^n) w(\mu)
d \mu$ \cite{Grunwald05}.  The corresponding estimator is the {\em
Bayesian posterior predictive distribution\/}
$P_{\text{Bayes}}(z_{i+1} \mid z^i) := \int M_{\mu}(z_{i+1}) w(\mu
\mid z^i) d \mu$ \cite{Grunwald05}. The Bayesian predictive
distribution is a mixture of elements of $\model$. We will call
standard estimators like the ML estimator, which are required to lie
in $\model$, {\em in-model\/} estimators. Estimators like the Bayesian
predictive distribution will be called {\em out-model}.

Let now $\breve{P}_n$ be any estimator, in-model or out-model. Let
$\breve{P}_{z^n}$ be the distribution estimated for a particular
realized sample $z^n$. We can measure the closeness of
$\breve{P}_{z^n}$ to $M_{\mu^*}$, the distribution in $\model$ closest
to $P$ in KL-divergence, by considering the {\em extended KL
divergence\/}
$$
D^*( M_{\mu^*} \| \breve{P}_{z^n} ) = E_{Z \sim P} [ - \ln \breve{P}_{z^n}(Z) 
 - [- \ln M_{\mu^*}(Z) ]].
$$
 We can now consider the {\em
  expected\/} KL divergence between $M_{\mu^*}$ and $\breve{P}_n$
after observing a sample of length $n$:
\begin{equation}
\label{eq:ekl}
E_{Z_1, \ldots, Z_n \sim P} [D^*( M_{\mu^*} \|
\breve{P}_n ) ].
\end{equation}
In analogy to the definition of `ordinary' {\em KL risk\/}
\cite{BarronRY98}, we call (\ref{eq:ekl}) the {\em extended KL risk}.
We recognize $\redundancy_U(n)$, the redundancy of the prequential
ML model, as the accumulated expected KL risk of our modified ML
estimator (see Proposition~\ref{prop:divbase} and
Lemma~\ref{lem:redundancy}).  In exactly the same way as for the
prequential ML code, the redundancy of the Bayesian code can be
re-interpreted as the accumulated KL risk of the Bayesian predictive
distribution.  With this interpretation, our Theorem~\ref{thm:main}
expresses that under misspecification, the cumulative KL risk of the
ML estimator differs from the cumulative KL risk of the Bayes
estimator by a term of $O(\ln n)$. If our conjecture that {\em no\/}
in-model estimator can achieve redundancy ${1 \over 2} \ln n + O(1)$
for all $\mu^*$ and all $P$ with finite variance is true
(Section~\ref{sec:other}), then it follows that the KL risk for
in-model estimators behaves in a fundamentally different way from the
KL risk for out-model estimators, and that out-model estimators are
needed to achieve the optimal constant $c= 1$ in the redundancy ${1
\over 2} c\ln n + O(1)$.

\section{Proof of Theorem~\ref{thm:main}}
\label{sec:proof}
\paragraph{Preliminaries}
Note that, for any $M_{\mu}, M_{\mu'} \in \model$, we have 
\begin{eqnarray*}
\label{eq:xnotz}
E_{P}[- \ln M_{\mu}(Z)] - E_{P}[- \ln M_{\mu'}(Z)]& = &
\eta(\mu)E_P[X(Z)] + \ln \partf(\eta(\mu)) +
E_{P}[- \ln h(Z)] \\ 
& - &\eta(\mu')E_P[X(Z)]
- \ln \partf(\eta(\mu')) - E_{P}[- \ln h(Z)]\\
& = &E_{P}[- \ln M_{\mu}(X)] - E_{P}[- \ln M_{\mu'}(X)],
\end{eqnarray*}
so that we have

\begin{proposition}
\label{prop:xmean}
  \begin{displaymath}
    \redundancy_U(n)=E_P[-\ln M(X^n)]-\inf_\mu E_P[-\ln\mmu(X^n)].
  \end{displaymath}
\end{proposition}
Proposition~\ref{prop:xmean} shows that relative redundancy, which is
the sole quantity of interest in the proof, depends only on the value
of $X$, not $Z$. Thus, in the proof of Theorem~\ref{thm:main} as well
as all the Lemmas and Propositions it makes use of, we will never
mention $Z$ again. Whenever we refer to a `distribution' we mean a
distribution of random variable $X$, and we also think of the data
generating distribution $P$ in terms of the distribution it induces on
$X$ rather than $Z$. Whenever we say `the mean' without further
qualification, we refer to the mean of the random variable $X$.
Whenever we refer to the Kullback-Leibler (KL) divergence between $P$
and $Q$, we refer to the KL divergence between the distributions they
induce for $X$ (the reader who is confused by this may simply restrict
attention to exponential family models for which $Z = X$, and consider
$X$ and $Z$ identical).

The proof refers to a number of theorems and lemmas which will be
developed in Section~\ref{section:building_blocks}. In the statement
of all these results, we assume, as in the statement of
Theorem~\ref{thm:main}, that $X, X_1, X_2, \ldots$ are i.i.d. $\sim P$
and that $\mu^*$ is the mean of $X$ under $P$.  If $X$ takes its
values in a countable set, then all integrals in the proof should be
read as the corresponding sums.

\begin{proof}{ \bf (of Theorem~\ref{thm:main})}
From Lemma~\ref{lem:redundancy} we have:
\begin{equation}
\redundancy_U(n)=\sum_{i=0}^{n-1}\,\mathop E_{\mihat\sim
  P}\left[D(\mmstar\|\mmihat)\right]
\label{eqn:redundancy}\end{equation}
Here, $\mihat$ is a random variable that takes on values according to
$P$, while $\mstar$ is fixed. We first abbreviate $\di=\mihat-\mstar$ and 
${d^k\over
d\mu^k}D(\mmstar\|\mmu)=\expdif{k}(\mu)$, That is, $\expdif{k}(\mu)$ is the $k$-th derivative of the function $f(\mu) := D(\mmstar\|\mmu)$. We now
Taylor-expand the divergence around $\mstar$:
$$D(\mmstar\|\mmihat)=0+\di\expdif{1}(\mstar)+{\di^2\over2}\expdif{2}(\mstar)+
  {\di^3\over6}\expdif{3}(\mstar)+{\di^4\over24}\expdif{4}(\mu)
$$ The last term is the remainder term of the Taylor expansion, in
which $\mu\in[\mstar,\mihat]$. The second term $\expdif{1}(\mstar)$ is
also zero, since $D(\mstar\|\mu)$ has its minimum at $\mu=\mstar$.
Now we rewrite:
$$\expdif{2}(\mu)={d^2\over
  d\mu^2}E[\ln\mmstar(X)-\ln\mmu(X)]=-{d^2\over
  d\mu^2}E[\ln\mmu(X)],$$
which, evaluated at $\mstar$, resembles the Fisher information. Fisher
information is usually defined as $I(\theta):=E\left[({d\over
d\theta}\ln f(X\mid\theta))^2\right]$, but as is well known
\cite{kassvos}, for exponential families this is equal to 
$-{d^2\over d\theta^2}E\left[\ln f(X\mid\theta)\right]$, which matches
$\expdif{2}(\cdot)$ exactly. Combining this with (\ref{eq:efficient})
(Section~\ref{sec:main}), we obtain:

\begin{equation}\label{eqn:term}
  D(\mmstar\|\mmihat)={1\over2}\di^2/\var_\mmstar(X)
  +{1\over6}\di^3 \expdif{3}(\mstar)+{1\over24}\di^4 \expdif{4}(\mu)
\end{equation}
We plug this expression back into Equation~\ref{eqn:redundancy}, giving
\begin{equation}
  \redundancy_U(n)=
{1\over2\var_\mmstar(X)}\sum_{i=0}^{n-1} E_P\left[\di^2\right]
+ R(n),
\label{eqn:rebbi}
\end{equation}
where the remainder term $R(n)$ is given by 

\begin{equation}\label{eq:remainder}
R(n) = \sum_{i=0}^{n-1} \mathop E_{\hat{\mu}_i \sim
  M^*}\left[{1\over6}\di^3 \expdif{3}(\mstar)+{1\over24}\di^4
  \expdif{4}(\mu)\right] 
\end{equation} 
where $\mu$ and $\di$ are random variables depending on $\mihat$ and
$i$. In Lemma~\ref{lem:hard} we show that $R(n) = O(1)$, giving:

\begin{equation}
  \redundancy_U(n)=O(1)+{1\over2\var_\mmstar(X)}\sum_{i=0}^{n-1} E_P\left[(\mihat-\mstar)^2\right]
\label{eqn:red2}
\end{equation}
Note that $\mihat$ is almost the ML estimator. This suggests that each
term in the sum of (\ref{eqn:red2}) should be almost equal to the
variance of the ML estimator, which is $\var X/i$. Because of the
slight modification that we made to the estimator, we get a correction
term of $O((i+1)^{-2})$ as established in
Theorem~\ref{thm:devavsquare}. This theorem gives:

\begin{eqnarray}
   \sum_{i=0}^{n-1} E_P\left[(\mihat-\mstar)^2\right]& = &\sum_{i=0}^{n-1}
  O\left((i+1)^{-2}\right)+\var_P(X)\sum_{i=0}^{n-1}(i+1)^{-1}\nonumber\\ 
  & = &O(1)+\var_P(X)\ln n
\label{eqn:devavsquare}
\end{eqnarray}
The combination of $(\ref{eqn:red2})$ and $(\ref{eqn:devavsquare})$ 
completes the proof.
\end{proof}

\section{Building blocks of the proof}
\label{section:building_blocks}
The proof of Theorem~\ref{thm:main} is based on
Lemma~\ref{lem:redundancy} and Lemma~\ref{lem:hard}. These Lemmas are
stated and proved, respectively, in Section~\ref{sec:expproperties}
and~\ref{sec:hard}.  The proofs of Theorem~\ref{thm:main} and
Theorem~\ref{thm:regret}, as well as the proof of both Lemmas, are
based on a number of generally useful results about probabilities and
expectations of deviations between the average and the mean of a
random variable.  Below, we first, in Section~\ref{sec:deviation},
list these deviation-related results.

\subsection{Results about Deviations between Average and Mean}
\label{sec:deviation}
\begin{lemma}\label{lemma:devsumsquare}
  Let $X, X_1, X_2, \ldots$ be i.i.d. with mean $0$. 
Then we have $E\left[\left(\sum_{i=1}^n
  X_i\right)^2\right]=n\var(X)$.
\end{lemma}
\begin{proof}
  For $n=0$ the lemma is obviously true. Suppose it is true for some
  $n$. For brevity we write $s_n=\sum_{i=1}^n X_i$. Because the mean
  is zero, we have $E\left[s_n\right]=\sum EX = 0$. Now we compute
  $E\left[s_{n+1}^2\right]=E\left[\left(s_n+X\right)^2\right]=E\left[s_n^2\right]+2E\left[s_n\right]EX+E[X^2]=(n+1)\var(X)$.
  The proof follows by induction.
\end{proof}

\begin{theorem}\label{thm:devavsquare}
Let $X, X_1, \ldots$ be i.i.d. random variables, define
$\mnhat:=(n_0\cdot x_0+\sum_{i=1}^n X_i)/(n+n_0)$ and $\mstar=E[X]$.
If $\var X < \infty$, then $E\left[(\mnhat-\mstar)^2\right]=
O\left((n+1)^{-2}\right)+\var(X)/(n+1)$.
\end{theorem}
\begin{proof}
We define $Y_i:=X_i-\mstar$; this can be seen as a new sequence of
i.i.d. random variables with mean $0$ and $\var Y=\var X$. We also set
$y_0:=x_0-\mstar$. Now we have:
\begin{eqnarray*}
  E \left[(\mnhat-\mstar)^2\right] & = &E
  \left[\left(n_0\cdot y_0+\sum_{i=1}^n Y_i\right)^2\right](n+n_0)^{-2}\\
  & = &E
  \Bigg[(n_0\cdot y_0)^2+2n_0\cdot y_0\sum_{i=1}^n Y_i+\left(\sum_{i=1}^n Y_i\right)^2\Bigg](n+n_0)^{-2}\\
  & = &O\left((n+1)^{-2}\right)+E
  \left[\left(\sum_{i=1}^n Y_i\right)^2\right](n+n_0)^{-2}\\
  & \overset{(*)} = &O\left((n+1)^{-2}\right)+n\var(Y)(n+n_0)^{-2}\\ & =
  &O\left((n+1)^{-2}\right)+\var(X)/(n+1),
\end{eqnarray*}
where $(*)$ follows by Lemma~\ref{lemma:devsumsquare}.
\end{proof}
The following theorem is of some independent interest.

\begin{theorem}
  \label{thm:devsumasym}
Suppose $X, X_1, X_2, \ldots $ are i.i.d. with mean 0.
  If the first $k\in\nats$ moments of $X$ exist, then we have 
Then $E\left[\left(\sum_{i=1}^n
  X_i\right)^k\right]=O\left(n^\floor{k\over2}\right)$.
\end{theorem}
\paragraph{Remark} 
It follows as a special case of Theorem~2 of \cite{Whittle1960} that
$E\left[|\sum_{i=1}^n X_i|^k\right]=O(n^{k\over2})$ which almost
proves this lemma and which would in fact be sufficient for our
purposes. We use this lemma instead which has an elementary proof.
\begin{proof}
We have:
  \begin{displaymath}
    E\!\left[\left(\sum_{i=1}^n
    X_i\right)^k\right]=E\!\left[\sum_{i_1=1}^n\cdots\sum_{i_k=1}^n
    X_{i_1}\cdots X_{i_k}\right]=\sum_{i_1=1}^n\cdots\sum_{i_k=1}^n E\!\left[
    X_{i_1}\cdots X_{i_k}\right]
  \end{displaymath}
  We define the \emph{frequency sequence} of a term to be the sequence
  of exponents of the different random variables in the term, in
  decreasing order. For a frequency sequence $f_1,\ldots,f_m$, we have
  $\sum_{i=1}^m f_i = k$. Furthermore, using independence of the
  different random variables, we can rewrite $E[X_{i_1}\cdots
  X_{i_k}]=\prod_{i=1}^m E[X^{f_i}]$ so the value of each term is
  determined by its frequency sequence. By computing the number of
  terms that share a particular frequency sequence, we obtain:
  \begin{displaymath}
    E\left[\left(\sum_{i=1}^n X_i\right)^k\right]=\sum_{f_1+\ldots+f_m=k}{n\choose m}{k\choose f_1,\ldots,f_m}\prod_{i=1}^m E[X^{f_i}]
  \end{displaymath}
  To determine the asymptotic behavior, first observe that the
  frequency sequence $f_1,\ldots,f_m$ of which the contribution grows
  fastest in $n$ is the longest sequence, since for that sequence the
  value of ${n\choose m}$ is maximized as
  $n\rightarrow\infty$. However, since the mean is zero, we can
  discard all sequences with an element $1$, because the for those
  sequences we have $\prod_{i=1}^m E[X^{f_i}]=0$ so they contribute
  nothing to the expectation. Under this constraint, we obtain the
  longest sequence for even $k$ by setting $f_i=2$ for all $1\le i\le
  m$; for odd $k$ by setting $f_1=3$ and $f_i=2$ for all $2\le i\le m$; in
  both cases we have $m=\floor{k\over2}$. The number of terms
  grows as ${n\choose m}\le n^m/m!=O(n^m)$; for
  $m=\floor{k\over2}$ we obtain the upper bound
  $O\left(n^\floor{k\over2}\right)$. The number of frequency sequences
  is finite and does not depend on $n$; since the contribution of each
  one is $O\left(n^\floor{k\over2}\right)$, so must be the sum.
\end{proof}

\begin{theorem}\label{thm:devavasym}
  Let $X, X_1, \ldots$ be i.i.d. random variables, define
  $\mnhat:=(n_0\cdot x_0+\sum_{i=1}^n X_i)/(n+n_0)$ and $\mstar=E[X]$.
  If the first $k$ moments of $X$ exist, then
  $E[(\mnhat-\mstar)^k]=O(n^{-\ceil{k\over2}})$.
\end{theorem}
\begin{proof}
  The proof is similar to the proof for Theorem~\ref{thm:devavsquare}.
  We define $Y_i:=X_i-\mstar$; this can be seen as a new sequence of
  i.i.d. random variables with mean $0$, and $y_0:=x_0-\mstar$. Now we
  have:
  \begin{eqnarray*}  
    E \left[(\mnhat-\mstar)^k\right]& = &E \left[\left(n_0\cdot
    y_0+\sum_{i=1}^n Y_i\right)^k\right](n+n_0)^{-k}\\
    & = &O\left(n^{-k}\right)\sum_{p=0}^k{k\choose p}(n_0\cdot y_0)^p
    E \left[\left(\sum_{i=1}^n Y_i\right)^{k-p}\right]\\
    & = &O\left(n^{-k}\right)\sum_{p=0}^k{k\choose p}(n_0\cdot y_0)^p\cdot
    O\left(n^{\floor{k-p\over2}}\right).
  \end{eqnarray*}
  In the last step we used Theorem~\ref{thm:devsumasym} to bound the
  expectation. We sum $k+1$ terms of which the term for $p=0$ grows
  fastest in $n$, so the expression is $O(n^{-\ceil{k\over2}})$ as
  required.
\end{proof}
Theorem~\ref{thm:devavasym}  concerns the \emph{expectation} of the
deviation of $\mnhat$. We also need a bound on the \emph{probability}
of large deviations. To do that we have the following separate
theorem:

\begin{theorem}\label{thm:boundprob}
Let $X, X_1, \ldots$ be i.i.d. random variables, define
  $\mnhat:=(n_0\cdot x_0+\sum_{i=1}^n X_i)/(n+n_0)$ and
  $\mstar=E[X]$. Let $k\in\{0,2,4,\ldots\}$. If the first $k$ moments
  exists then
  $P(|\mnhat-\mstar|\ge\delta)=O\left(n^{-\ceil{k\over2}}\delta^{-k}\right)$.
\end{theorem}
\begin{proof}
  \begin{eqnarray*}
    P(|\mnhat-\mstar|\ge\delta)& =
    &P\left((\mnhat-\mstar)^k\ge\delta^k\right)\\ & \le
    &E\left[(\mnhat-\mstar)^k\right]\delta^{-k}\quad\hbox{(by Markov's
    inequality)}\\ & =
    &O\left(n^{-{k\over2}}\delta^{-k}\right)\qquad\hbox{(by
    Theorem~\ref{thm:devavasym})}
  \end{eqnarray*}
\end{proof}

\subsection{Lemma~\ref{lem:redundancy}: Redundancy for Exponential Families}
\label{sec:expproperties}
\begin{lemma}
Let $U$ be a prequential ML model and $\model$ be an
  exponential family as in Theorem~\ref{thm:main}. We have
\begin{displaymath}
  \redundancy_U(n)=\sum_{i=0}^{n-1}\,\mathop E_{\mihat\sim
  P}\left[D(\mmstar\parallel \mmihat)\right].
\end{displaymath}
(Here, the notation $\mihat\sim P$ means that we take the
expectation with respect to $P$ over data sequences of length $i$,
of which $\mihat$ is a function.)
\label{lem:redundancy}
\end{lemma}
\begin{proof} 
 We have:
 \begin{displaymath}
   \arg\inf_\mu E_P\left[-\ln \mmu(X^n)\right]=\arg\inf_\mu E_P\left[\ln{\mmstar(X^n)\over\mmu(X^n)}\right]=\arg\inf_\mu D(\mmstar\parallel\mmu)
 \end{displaymath}
 In the last step we used Proposition~\ref{prop:divbase} below. The
 divergence is minimized when $\mu=\mstar$ \cite{kassvos}, 
so we find that:
  \begin{multline}
    \redundancy_U(n) = E_P[-\ln U(X^n)]-E_P[-\ln\mmstar(X^n)]
     = E_P\left[\ln{\mmstar(X^n)\over U(X^n)}\right]\\
     = E_P\left[\sum_{i=0}^{n-1}\ln{\mmstar(X_i)\over\mmihat(X_i)}\right]
     = \sum_{i=0}^{n-1} E_P\left[\ln{\mmstar(X_i)\over\mmihat(X_i)}\right]
     = \sum_{i=0}^{n-1} \mathop{E}_{\mihat\sim P}\left[D(\mmstar\parallel\mmihat)\right].
  \end{multline}
  Here, the last step again follows from Proposition~\ref{prop:divbase}.
\end{proof}
\begin{proposition}\label{prop:divbase}
Let $X \sim P$ with mean $\mu^*$, and let $M_{\mu}$ index an exponential family with sufficient statistic $X$, so that $M_{\mu^*}$ exists. We have:
\begin{displaymath}
  E_P\left[-\ln{M_{\mu^*}(X)\over M_\theta(X)}\right]=D(M_{\mu^*}
\parallel M_\theta)
\end{displaymath}
\end{proposition}
\begin{proof}
  Let $\eta(\cdot)$ denote the function mapping parameters in the mean
  value parameterization to the natural parameterization. (It is the
  inverse of the function $\mu(\cdot)$ which was introduced in the
  discussion of exponential families.) By working out both sides of
  the equation we find that they both reduce to:
  \begin{displaymath}
    \eta(\mu^*)\mu^*+\ln \partf(\eta(\mu^*))-\eta(\theta)\mu^*-\ln
    \partf(\eta(\theta)).
  \end{displaymath}
\end{proof}

\subsection{Lemma~\ref{lem:hard}: Convergence of the sum of the remainder terms}
\label{sec:hard}
\begin{lemma}
\label{lem:hard}
Let $R(n)$ be defined as in (\ref{eq:remainder}). Then
$$
R(n) = O(1).$$ 
\end{lemma}
\begin{proof}
We omit irrelevant constants and the term for the first outcome, which
is well-defined because of our modification of the ML estimator. We
abbreviate ${d^k\over d\mu^k}D(\mmstar\|\mmu)=\expdif{k}(\mu)$ as in
the proof of Theorem~\ref{thm:main}. First we consider the third order
term. We write $\mathop E_{\di\sim P}$ to indicate that we take the
expectation over data which is distributed according to $P$, of which
$\di$ is a function. We use Theorem~\ref{thm:devavasym} to bound the
expectation of $\di^3$; under the condition that the first three
moments exist, which is assumed to be the case, we obtain:
$$  \sum_{i=1}^{n-1}\mathop{E}_{\di\sim
  P}\left[\di^3
  \expdif{3}(\mstar)\right]=\expdif{3}(\mstar)\sum_{i=1}^{n-1}
E[\delta_i^3]= \expdif{3}(\mstar)\sum_{i=1}^{n-1}
  O(i^{-2})=
 O(1). 
$$
(The constants implicit in the big-ohs are the same across terms.)
%steven: removed unnecessary \multline in the above

The fourth order term is more involved, because $\expdif{4}(\mu)$ is not
necessarily constant across terms. To compute it we first distinguish
a number of regions in the value space of $\di$: let
$\Delta_-=(-\infty,0)$ and let $\Delta_0=[0,a)$ for some constant
value $a>0$. If the individual outcomes $X$ are bounded on the
right hand side by a value $g$ then we require that $a<g$ and we
define $\Delta_1=[a,g)$; otherwise we define $\Delta_j=[a+j-1, a+j)$
for $j\ge1$. Now we must establish convergence of:
\begin{displaymath}
\sum_{i=1}^{n-1} \mathop{E}_{\di\sim P}\left[\di^4
  \expdif{4}(\mu)\right]~=~\sum_{i=1}^{n-1}\sum_j
  P(\di\in\Delta_j)\mathop{E}_{\di\sim P}\left[\di^4
  \expdif{4}(\mu)\mid\di\in\Delta_j\right]
\end{displaymath}
If we can establish that the sum converges for all regions $\Delta_j$
for $j\ge0$, then we can use a symmetrical argument to establish
convergence for $\Delta_-$ as well, so it suffices if we restrict
ourselves to $j\ge0$. First we show convergence for $\Delta_0$. In
this case, the basic idea is that since the remainder $\expdif{4}(\mu)$ is
well-defined over the interval $\mstar\le\mu<\mstar+a$, we can bound
it by its extremum on that interval, namely
$m:=\sup_{\mu\in[\mstar,\mstar+a)}\left|\expdif{4}(\mu)\right|$. Now we get:
\begin{displaymath}
  \left|\sum_{i=1}^{n-1}P(\di\in\Delta_0)E\left[\di^4 \expdif{4}(\mu)\mid\di\in\Delta_0\right]\right|
  ~\le~\left|\sum_{i=1}^{n-1}1\cdot
  E\left[\di^4\left|\expdif{4}(\mu)\right|\right]\right|
  ~\le~\left|m\sum_i E\left[\di^4\right]\right|
\end{displaymath}
Using Theorem~\ref{thm:devavasym} we find that $E[\di^4]$ is
$O(i^{-2})$ of which the sum converges.
Theorem~\ref{thm:devavasym} requires that the first four moments of
$P$ exist, but this is guaranteed to be the case: either the outcomes are
bounded from both sides, in which case all moments necessarily exist,
or the existence of the required moments is part of the condition on
the main theorem.

Now we have to distinguish between the unbounded and bounded
cases. First we assume that the $X$ are unbounded from above. In
this case, we must show convergence of:
\begin{displaymath}
  \left|\sum_{i=1}^{n-1}\sum_{j=1}^\infty
  P(\di\in\Delta_j)E\left[\di^4 \expdif{4}(\mu)\mid\di\in\Delta_j\right]\right|
\end{displaymath}
We bound this expression from above. The $\di$ in the expectation is
at most $a+j$. Furthermore $\expdif{4}(\mu)=O(\mu^{k-6})$ by assumption on
the main theorem, where $\mu\in[a+j-1,a+j)$. Depending on $k$, both
boundaries could maximize this function, but it is easy to check that
in both cases the resulting function is $O(j^{k-6})$. So we get:
\begin{displaymath}
  \ldots \le
  \sum_{i=1}^{n-1}\sum_{j=1}^\infty P(\left|\di\right|\ge
  a+j-1)(a+j)^4 O(j^{k-6})
\end{displaymath}
Since we know from the condition on the main theorem that the first
$k\ge4$ moments exist, we can apply Theorem~\ref{thm:boundprob} 
to find that $P(|\di|\ge
a+j-1)=O(i^{-\ceil{k\over2}}(a+j-1)^{-k})=O(i^{-{k\over2}})O(j^{-k})$
(since $k$ has to be even); plugging this into the equation and
simplifying we obtain $\sum_i O(i^{-{k\over2}})\sum_j O(j^{-2})$. For
$k\ge4$ this expression converges.

Now we consider the case where the outcomes are bounded from above by
$g$.  This case is more complicated, since now we have made no extra
assumptions as to existence of the moments of $P$. Of course, if the
outcomes are bounded from both sides, then all moments necessarily
exist, but if the outcomes are unbounded from below this may not be
true. We use a trick to remedy this: we map all outcomes into a new
domain in such a way that all moments of the transformed variables are
guaranteed to exist. Any constant $x^-$ defines a mapping
$g(x):=\max\{x^-,x\}$. Furthermore we define the random variables
$Y_i:=g(X_i)$, the initial outcome $y_0:=g(x_0)$ and the mapped
analogues of  $\mstar$ and $\mihat$, respectively: $\mdagger$ is defined
as the mean of $Y$ under $P$ and $\mitilde:=(y_0\cdot n_0+\sum_{j=1}^i
Y_j)/(i+n_0)$. Since $\mitilde\ge\mihat$, we can bound:
\begin{eqnarray*}
  \left|\sum_i P(\di\in\Delta_1)E\left[\di^4
  \expdif{4}(\mu)\mid\di\in\Delta_1\right]\right|
  & \le &\sum_i P(\mihat-\mstar\ge a)\sup_{\di\in\Delta_1}\left|\di^4
  \expdif{4}(\mu)\right|\\
  & \le &\sum_i P(|\mitilde-\mdagger|\ge
  a+\mstar-\mdagger)g^4\sup_{\di\in\Delta_1}\left|
  \expdif{4}(\mu)\right|%\\
% & \le &\sum_i
%  O\left(i^{-{k\over2}}\right)g^4\sup\left|\expdif{4}(\mu)\right|
\end{eqnarray*}
By choosing $x^-$ small enough, we can bring $\mdagger$ and $\mstar$
arbitrarily close together; in particular we can choose $x^-$ such
that $a+\mstar-\mdagger>0$ so that application of
Theorem~\ref{thm:boundprob} is safe. It reveals that the summed
probability is $O(i^{-{k\over2}})$. Now we bound $\expdif{4}(\mu)$
which is $O((g-\mu)^{-m})$ for some $m\in\nats$ by the condition on
the main theorem. Here we use that $\mu\le\mihat$; the latter is
maximized if all outcomes equal the bound $g$, in which case the
estimator equals $g-n_0(g-x_0)/(i+n_0)=g-O(i^{-1})$. Putting all of
this together, we get
$\sup\left|\expdif{4}(\mu)\right|=O((g-\mu)^{-m})=O(i^m)$; if we plug
this into the equation we obtain:
\begin{displaymath}
  \ldots~\le~\sum_i O(i^{-{k\over2}})g^4O(i^m)=g^4\sum_i O(i^{m-{k\over2}})
\end{displaymath}
This converges if we choose $k\ge6m$. We can do this because the
construction of the mapping $g(\cdot)$ ensures that all moments exist,
and therefore certainly the first $6m$.
\end{proof}

\section{Proof of Theorem~\ref{thm:regret}}
\label{sec:proofb}
We use the same conventions as in the proof of Theorem~\ref{thm:main}.
Specifically, we concentrate on the random variables $X_1, X_2,
\ldots$ rather than $Z_1, Z_2, \ldots$, which is justified by
Equation~(\ref{eq:xnotz}). Let $f(x^n) = - \ln M_{\mu^*}(x^n) - [
\inf_{\mu \in {\Theta_\mu} } - \ln M_{\mu}(x^n)]$. Within this
section, $\hat{\mu}(x^n)$ is defined as the ordinary ML estimator.
Note that, if $x^n$ is such that its ML estimate is defined, then
$f(x^n) = - \ln M_{\mu^*}(x^n) + \ln M_{\hat{\mu}(x^n)}(x^n)$.

Note $d(n) = E_P[f(X^n)]$. Let $h(x)$ be the carrier of the
exponential family under consideration (see
Definition~\ref{def:expfam}). Without loss of generality, we assume
$h(x) > 0$ for all $x$ in the finite set ${\cal X}$. Let $a_n^2 =
n^{-1/2}$. We can write
\begin{eqnarray}
\label{eq:sylvester}
d(n) ~=~ E_P [f(X^n)] ~=~ 
& \pi_n  & E_P [f(X^n) \mid (\mu^* - \hat{\mu}_n)^2 \geq
a_n^2 ] \nonumber \\
+ & (1- \pi_n)&E_P [f(X^n) \mid (\mu^* - \hat{\mu}_n)^2 <
a_n^2 ], 
\end{eqnarray}
where $\pi_n = P( (\mu^* - \hat{\mu}_n)^2 \geq a_n^2)$.  We determine
$d(n)$ by bounding the two terms on the right of
(\ref{eq:sylvester}). We start with the first term.  Since $X$ is
bounded, all moments of $X$ exists under $P$, so we can bound $\pi_n$
using Theorem~\ref{thm:boundprob} with $k = 8$ and $\delta =
a_n = n^{-1/4}$. (Note that the theorem in turn makes use of
Theorem~\ref{thm:devavasym} which remains valid when we use $n_0=0$.)
This gives
\begin{equation}
\label{eq:vuurwerk}
\pi_n = O(n^{-2}).
\end{equation}
Note that for all $x^n \in \xspace^n$, we have
\begin{equation}
\label{eq:knal}
0 \leq  f(x^n) \leq \sup_{x^n \in \xspace^n} f(x^n) \leq 
\sup_{x^n \in \xspace^n} - \ln M_{\mu^*}(x^n)
\leq  n C,
\end{equation}
where $C$ is some constant. Here the first inequality follows because
$\hat{\mu}$ maximizes $\ln M_{\hat{\mu}(x^n)}(x^n)$ over $\mu$;
the second is immediate; the third follows because we are dealing with
discrete data, so that $M_{\hat{\mu}}$ is a probability mass
function, and $M_{\hat{\mu}}(x^n)$ must be $\leq 1$. The final
inequality follows because $\mu^*$ is in the interior of the parameter
space, so that the natural parameter $\eta(\mu^*)$ is in the interior
of the natural parameter space. Because $X$ is bounded and we assumed $h(x) > 0$ for all $x \in \xspace$, it follows by
the definition of exponential families that $\sup_{x \in \xspace}
- \ln M_{\mu^*}(x) < \infty$.

Together (\ref{eq:vuurwerk}) and (\ref{eq:knal}) show that the
expression on the first line of (\ref{eq:sylvester}) converges to $0$,
so that (\ref{eq:sylvester}) reduces to
\begin{equation}
\label{eq:trein}
d(n) = (1- \pi_n)  E_P [f(X^n) \mid (\mu^* - \hat{\mu}_n)^2 <
a_n^2 ] + O(n^{-1}).
\end{equation}
To evaluate the term inside the expectation 
further we first Taylor approximate $f(x^n)$ 
around $\hat{\mu}_n = \hat{\mu}(x^n)$, for given $x^n$
with $(\mu^* - \hat{\mu}_n)^2 <
a_n^2 = 1 /\sqrt{n}$. We get

\begin{equation}
\label{eq:sailor}
f(x^n) = 
- (\mu^* - \hat{\mu}_n) \frac{d}{d \mu} \ln M_{\hat{\mu}_n}(x^n)
+ n \frac{1}{2} (\mu^* - \hat{\mu}_n)^2 I(\mu_n),
\end{equation}
where $I$ is the Fisher information (as defined in
Section~\ref{sec:proof}) and $\mu_n$ lies in between $\mu^*$ and
$\hat{\mu}$, and depends on the data $x^n$. 
Since the first derivative of $\mu$ at the ML estimate
$\hat{\mu}$ is $0$, the first-order term is 0. Therefore
$
f(x^n) = \frac{1}{2} n (\mu^* - \hat{\mu}_n)^2 I(\mu_n),
$
so that 
\begin{displaymath}
\label{eq:jaja}
\frac{1}{2} n g(n) \!\!\!\inf_{\mu \in [\mstar - a_n, \mstar + a_n]}\!\!\! I(\mu)
\leq E_P [f(X^n) \mid (\mstar - \mnhat)^2 <
a_n^2 ] \leq 
\frac{1}{2} n  g(n) \!\!\!\sup_{\mu \in [\mstar - a_n, \mstar + a_n]}\!\!\!
I(\mu),
\nonumber
\end{displaymath}
where we abbreviated
$g(n) := E_P [ (\mu^* - \hat{\mu}_n)^2 \mid (\mu^* - \hat{\mu}_n)^2 <
a_n^2 ].$
Since $I(\mu)$ is smooth and positive, we can Taylor-approximate it as
$I(\mstar)+O(n^{-{1\over4}})$, so we obtain the bound:
\begin{equation}\label{eqn:fisherbound}
  E_P [f(X^n) \mid (\mstar - \mnhat)^2 <a_n^2]= ng(n)\left({1\over2}I(\mstar)+O(n^{-{1\over4}})\right).
\end{equation}
To evaluate $g(n)$, note that we have
\begin{equation}
E_P [ (\mstar - \mnhat)^2 ] = 
\pi_n E_P [ (\mstar - \mnhat)^2 \mid (\mstar - \mnhat)^2 \geq
a_n^2 ] +(1- \pi_n) g(n).
\end{equation} 
Using Theorem~\ref{thm:devavsquare} with $n_0=0$ we rewrite the
expectation on the left hand side as $\var_P X/n$. Subsequently
reordering terms we obtain:
\begin{equation}\label{eq:bus}
g(n) = \frac{(\var_P X)/ n - \pi_n E_P [ (\mstar - \mnhat)^2 \mid (\mstar - \mnhat)^2 \geq a_n^2 ]}{1- \pi_n}.
\end{equation}
Plugging this into bound~(\ref{eqn:fisherbound}), and multiplying both sides by
$1- \pi_n$, we get:
\begin{multline}
(1- \pi_n) E_P [f(X^n) \mid (\mu^* - \hat{\mu}_n)^2 <
a_n^2 ] = \\ 
\left( \var_P X - n \pi_n 
E_P [ (\mu^* - \hat{\mu}_n)^2 \mid (\mu^* - \hat{\mu}_n)^2 \geq
a_n^2 ] \right) \left( \frac{1}{2}I(\mu^*) + O(n^{-{1\over4}})\right).
\nonumber
\end{multline}
Since $X$ is bounded, the expectation on the right must lie between
$0$ and some constant $C$. Using $\pi_n = O(n^{-2})$ and the fact that
$I(\mu^*) = 1/ \var_{M_{\mu^*}} X$ (Equation (\ref{eq:efficient})), we
get
$$(1- \pi_n) E_P [f(X^n) \mid (\mu^* - \hat{\mu}_n)^2 < a_n^2 ] =
\frac{1}{2} \frac{\var_P X}{\var_{M_{\mu^*}} X} +
O(n^{-{1\over4}}).$$
The result follows if we combine this with (\ref{eq:trein}).

\section{Conclusion and Future Work}
In this paper we established two theorems about the relative redundancy, defined in Section~\ref{sec:introres}:
\begin{enumerate}
\item A particular type of universal code, the \emph{prequential ML
  code} or \emph{ML plug-in code}, exhibits behavior that we found
  unexpected. While other important universal codes such as the
  NML/Shtarkov and Bayesian codes, achieve a regret of ${1\over2}\ln
  n$, where $n$ is the sample size, the prequential ML code achieves
  a relative redundancy of ${1\over2}{\var_P X\over\var\mmstar X}\ln
  n$. (Sections~\ref{sec:introres} and \ref{sec:main}.)
\item At least for finite sample spaces, the relative redundancy is
  very close to the expected regret, the difference going to
  ${1\over2}{\var_P X\over\var_\mmstar X}$ as the sample size
  increases (Section~\ref{sec:regret}, Theorem~\ref{thm:regret}).
In future work, we hope to extend this theorem to general 
1-parameter exponential families with arbitrary sample spaces.
\end{enumerate}
Under the heading ``Related Work'' in Section~\ref{sec:introres} we
list a substantial amount of literature in which the regret for the
prequential ML code is proven to grow with ${1\over2}\ln n$. While
this may seem to contradict our results, in fact it does not: In
those articles, settings are considered where $P\in\model$, and
under such circumstances our own findings predict precisely that
behavior.

The first result is robust with respect to slight variations in the
definition of the prequential ML code: in our framework the so-called
``start-up problem'' (the unavailability of an ML estimate for the
first few outcomes) is resolved by introducing fake initial
outcomes. Our framework thus also covers prequential codes that use
other point estimators such as the Bayesian MAP and mean estimators
defined relative to a large class of reasonable priors. In
Section~\ref{sec:other} we conjecture that no matter what in-model
estimator is used, the prequential model cannot yield a relative
redundancy of ${1\over2}\ln n$ independently of the variance of the
data generating distribution.

\section{Acknowledgment}
This work was supported in part by the IST Programme of the European
Community, under the PASCAL Network of Excellence,
IST-2002-506778. This publication only reflects the authors' views.

\bibliography{master,MDL,prequential}
\end{document}